\documentclass[journal]{IEEEtran}

\ifCLASSINFOpdf
\else
   \usepackage[dvips]{graphicx}
\fi
\usepackage{url}

\usepackage{graphicx}

\usepackage{color}
\usepackage{amsmath}
\usepackage{enumerate}
\usepackage{cite}
\usepackage{subfigure}
\usepackage{flushend}
\usepackage{booktabs}
\usepackage{bm}
\usepackage{algorithm}
\usepackage{algpseudocode}
\usepackage{amssymb}
\usepackage{mathtools}
\usepackage{breqn}
\usepackage{threeparttable}

\usepackage[OT1]{fontenc}

\begin{document}
	\title{Competition-based Adaptive ReLU \\ for Deep Neural Networks}
	\author{Junjia~Chen~and~Zhibin~Pan
	\thanks{The authors are with the School of Electronic and Information Engineering, Xi’an Jiaotong University, Xi’an 710049, P. R. China. (e-mail: zbpan@xjtu.edu.cn)}}


	\maketitle
	
	\begin{abstract}
		Activation functions introduce nonlinearity into deep neural networks. Most popular activation functions allow positive values to pass through while blocking or suppressing negative values. From the idea that positive values and negative values are equally important, and they must compete for activation, we proposed a new Competition-based Adaptive ReLU (CAReLU). CAReLU scales the input values based on the competition results between positive values and negative values. It defines two parameters to adjust the scaling strategy and can be trained uniformly with other network parameters. We verify the effectiveness of CAReLU on image classification, super-resolution, and natural language processing tasks. In the experiment, our method performs better than other widely used activation functions. In the case of replacing ReLU in ResNet-18 with our proposed activation function, it improves the classification accuracy on the CIFAR-100 dataset. The effectiveness and the new perspective on the utilization of competition results between positive values and negative values make CAReLU a promising activation function.
	\end{abstract}
	
	
	\IEEEpeerreviewmaketitle
	
	\section{Introduction}
	\IEEEPARstart{D}{eep} learning is one of the most popular techniques demonstrating superior performance across multiple applications and has been the focus of academic and engineering communities for years. Deep learning techniques are evolving fast and each year hundreds of new models are proposed. Its applications quickly dominate in many areas such as computer vision (CV)\cite{nn_gpgpu_3, misc_transformer} and natural language processing (NLP)\cite{misc_attention, dl_nlp_overview}.


	\begin{equation} \label{nn_node_op}
		y = g(\bm{w}^{\rm T} \bm{x} + b).
	\end{equation}

	A deep neural network contains thousands of neural units. Every neural unit gathers the input signals from other neurons and generates a signal for other neurons to process. The structure of a deep neural network is a directed acyclic graph where every vertex represents a mapping from signals of incoming arcs into new signals of outgoing arcs. Eq. \eqref{nn_node_op} is one of the most common vertices in deep neural networks, which applies an affine transformation and a non-linear mapping $g(\cdot)$ to the input tensor, where $\bm{x} \in \mathbb{R}^{d} $ is the $d$-dimension input gathered from incoming arcs, $\bm{w} \in \mathbb{R}^{d}$ is the weight vector, $b \in \mathbb{R}$ is the bias, and $y \in \mathbb{R}$ is the vertex's output. $\bm{w}$ and $b$ can be obtained through training. This non-linear mapping is called the activation function, without which a deep neural network degenerates into a linear regression model.

	\begin{equation} \label{relu_formular}
		{\rm ReLU}(\bm{w}^{\rm T} \bm{x} + b) = {\rm max}(\bm{w}^{\rm T} \bm{x} + b, 0).
	\end{equation}

	In the early stage of deep learning research, S-shape functions like the Sigmoid function and the hyperbolic tangent function (tanh) were proposed and became popular but they were later found ineffective. Rectified Linear Unit (ReLU)\cite{relu_1,relu_2} shown in Eq. \eqref{relu_formular} was proposed that demonstrates top-tier convergence speed, better generalization and ease of implementation. But ReLU completely shuts down negative values, which led to the development of LeakyReLU\cite{leakyrelu} that allows the negative values to pass through. Parametric ReLU (PReLU)\cite{prelu} improves model fitting by parameterizing the LeakyReLU's slope of the negative part. Multi-phase ReLU\cite{multiphaserelu} utilizes six different phases from the input. Sparse regularization\cite{sparsereg} increases the sparsity of ReLU's input.

	Activation functions with more complicated formulas are also explored. Dan proposed Gaussian Error Linear Unit (GeLU) and Sigmoid Linear Unit (SiLU)\cite{gelu}, which are widely used in natural language processing. SiLU was also found and named Swish\cite{swish} by Prajit by utilizing meta-learning techniques. Inspired by SiLU's self-gating property, Mish\cite{mish} and Serf\cite{serf} are proposed to further improve performance. Brosnan developed the Universal Activation Function (UAV) \cite{uav}, which generalizes some popular activation functions by using five trainable parameters. Shui-Long proposed the tanhLU\cite{tanhlu} by integrating tanh into a linear unit. Mathew proposed the Growing Cosine Unit (GCU)\cite{growingcos} that can improve gradient flow and reduce network size.

	Most activation functions differently process the negative part and the positive part of the input tensor. For example, ReLU completely blocks negative values; LeakyReLU, PReLU, and SiLU scale down negative values. It implies that positive values are far more important than negative values. From the viewpoint of signal processing, positive and negative values are equally important and they must compete for activation instead of simply shutting down or suppressing negative values.

	In this paper, we propose a new Competition-based Adaptive ReLU (CAReLU). 1) A CAReLU utilizes the competition results between positive values and negative values in the input tensor. 2) A CAReLU has two parameters that can be trained uniformly with other network parameters. 3) Different competition indicators can cooperate with CAReLU to generate different activation functions for different tasks. We evaluate our method and find that our method constantly performs better compared to the most popular activation functions.
	
	\section{Proposed Method}

	\subsection{Construction of Competition-based Adaptive ReLU}


	Our proposed method stems from the idea that both positive values and negative values are equally important and they must compete for activation. We choose energy as an indicator of the competition. The rule is that the competitor with higher energy wins the qualification for activation. Let $\bm{z} = (z_1, z_2, ..., z_d) \in \mathbb{R}^d$ be the output of the previous affine transformation. The percentage of positive values' energy $p_E$ is defined as follows:
	\begin{equation} \label{p_E}
		\begin{aligned}
			p_E &= \frac{\sum_{j=1}^{d} [{\rm max}(z_j, 0)]^2}{\Vert \bm{z} \Vert^2 + \epsilon}, \\
		\end{aligned}
	\end{equation}
	where a small positive constant value $\epsilon$ is added in the denominator to prevent the division by zero. We define $s$ as follows:
	\begin{equation} \label{SDF1}
		\begin{aligned}
			s &= {\rm sgn}(2p_E-1), \\
		\end{aligned}
	\end{equation}
	where ${\rm sgn}(\cdot)$ is the sign function. If $p_E < 0.5$ is true, then we have $s=-1$, which means the negative values win the competition; If $p_E > 0.5$ is true, then we have $s=1$, which means the positive values win the competition. By multiplying $\bm{z}$ with $s$ when passing it to the ReLU, we can flip the input if negative values have higher energy:
	\begin{equation} \label{inputflip}
		y_i = {\rm ReLU}(s z_i).
	\end{equation}

	However, this vanilla version of our idea does not perform well. First, there is no degeneracy into regular ReLU when prioritizing higher energy impairs performance. Second, the sign function creates discontinuities in the model's parameter space, which harms the gradient-based optimization. To address the first issue, we replace the fixed scaling factor of $2$ and the bias of $-1$ in Eq. \eqref{SDF1} with trainable parameters $\alpha$ and $\beta$ per layer:
	\begin{equation} \label{SDF2}
		\begin{aligned}
			s &= {\rm sgn}(\alpha p_E + \beta). \\
		\end{aligned}
	\end{equation}
	If $\alpha$ is close or equal to $0$ after training, this mapping degenerates into ReLU with a factor of $1$ or $-1$. Only 2 extra parameters are introduced per layer, which is negligible when considering the total number of weights.



	To address the second issue, we replace the sign function with the tanh function as follows:
	\begin{equation} \label{SDF3}
		\begin{aligned}
			s &= {\tanh}(\alpha p_E + \beta). \\
		\end{aligned}
	\end{equation}
	The tanh function has a codomain ranging from $-1$ to $1$ and it is a continuous function with a non-zero gradient, which can be viewed as a smooth version of the sign function.
	
	There are other kinds of indicators for competition between positive values and negative values other than energy. The other two indicators we experiment with are: 
	\begin{equation} \label{p_{L1}}
		\begin{aligned}
			p_{L1} &= \frac{\sum_{j=1}^{d} \vert {\rm max}(z_j, 0) \vert}{\Vert \bm{z} \Vert_1 + \epsilon} \\
		\end{aligned},
	\end{equation}
	\begin{equation} \label{p_cnt}
		\begin{aligned}
			p_c &= \frac{\sum_{j=1}^{d} H(z_j)}{d} \\
		\end{aligned},
	\end{equation}
	where $H(\cdot)$ is the Heaviside step function, Eq. \eqref{p_{L1}} is the percentage of positive values' L1-norm, and Eq. \eqref{p_cnt} is the percentage of the number of positive values. 

	Competition-based Adaptive Scaling (CAS) can be defined as follows to adaptively scale the input $\bm{z}$ based on competition results:
	\begin{equation} \label{CAS}
			{\rm CAS}(\bm{z}) \triangleq K[{\tanh}(\alpha p + \beta)]\bm{z}, \\
	\end{equation}
	where $K$ is a constant and $p \in \{p_E, p_{L1}, p_c\}$. Since $|\tanh(\alpha p + \beta)|$ is less than 1, the magnitude of feature vectors will shrink layer by layer in a sequence model. To combat this phenomenon, the constant $K$ is placed to scale up the results of the tanh function. By concatenating CAS and ReLU, we construct Competition-based Adaptive ReLU as follows:
	\begin{equation} \label{CAReLU}
			{\rm CAReLU}(\bm{z}) \triangleq {\rm ReLU}({\rm CAS}(\bm{z})).
	\end{equation}
	
	\subsection{Working with Batch Normalization}
	Batch Normalization (BN) \cite{bn} is widely used in convolutional neural networks. BN normalizes and rescales the input with 2 parameters on a mini-batch before activation. Since CAReLU requires competition between positive values and negative values, the BN's normalization might impair this process. To improve our method's compatibility with BN, we define $\rm BN\text{-}CAReLU$ as follows by placing Eq. \eqref{CAS} before BN so that the neural network can obtain competition results before the normalization:
	\begin{equation} \label{CASBNReLU}
			{\rm BN\text{-}CAReLU}(\bm{z}) \triangleq {\rm ReLU}({\rm BN}({\rm CAS}(\bm{z}))).
	\end{equation}

	\subsection{Gradients of the Competition-based Adaptive Scaling} 
	Let $\alpha_0$ be the initial value of $\alpha$ and $\beta_0$ be the initial value of $\beta$. Instead of running a grid search for initial values, we simply set these values as follows:
	\begin{equation}
		\begin{dcases}
			\alpha_0 &= 0, \\
			\beta_0 &= 1, \\
			K &= 1/\tanh(\beta_0) \approx 1.313.
		\end{dcases}
	\end{equation}
	where $\alpha_0=0$ means that deep neural networks don't utilize competition results at the beginning. $\beta_0=1$ makes the tanh function starts at the non-saturating area, where the gradient is large enough to initiate the update. The constant $K$ is set to $1/\tanh(\beta_0)$ so that the CAS is initialized as an identity mapping when the training begins.

	To update the parameters, we need to compute gradients of the loss function with respect to $\alpha$, $\beta$, and input value $z_i$. Let $\hat{\bm{z}}=({\hat{z}}_1, {\hat{z}}_2, ..., {\hat{z}}_d) \in \mathbb{R}^{d}$ be the output of CAS as follows:
	\begin{equation}
		\hat{\bm{z}}={\rm CAS}(\bm{z}).
	\end{equation}
	Gradients can be derived from the chain rule as follows:
	\begin{dmath}
		\begin{aligned}
			\frac{\partial \mathcal{L}}{\partial \alpha} =& K\frac{4p}{({\rm e}^{\alpha p + \beta} + {\rm e}^{-\alpha p - \beta})^2}\sum_{j=1}^{d} \frac{\partial \mathcal{L}}{\partial \hat{z}_j} z_j, \\
			\frac{\partial \mathcal{L}}{\partial \beta} =& K\frac{4}{({\rm e}^{\alpha p + \beta} + {\rm e}^{-\alpha p - \beta})^2}\sum_{j=1}^{d} \frac{\partial \mathcal{L}}{\partial \hat{z}_j} z_j, \\
			\frac{\partial \mathcal{L}}{\partial z_i} =& K[\frac{4\alpha}{({\rm e}^{\alpha p + \beta} + {\rm e}^{-\alpha p - \beta})^2} \frac{\partial p}{\partial z_i} (\sum_{j=1}^{d} \frac{\partial \mathcal{L}}{\partial \hat{z}_j} z_j) \\
			&+ \frac{\partial \mathcal{L}}{\partial \hat{z}_i} \tanh(\alpha p + \beta)],
		\end{aligned}
	\end{dmath}
	where $\mathcal{L}$ is the loss function and $\frac{\partial \mathcal{L}}{\partial \hat{z}_i}$ is obtained from backpropagation. $\frac{\partial p}{\partial z_i}$ is the partial differential of the competition indicator $p$ with respect to the input. For $p_E$, $p_{L1}$ and $p_c$, we ignore the small positive constant value $\epsilon$ and the gradients are derived as follows:


	\begin{equation} \label{dp_e_dz_i}
		\begin{aligned}
			\frac{\partial p_E}{\partial z_i} &= \frac{2 {\rm max}(z_i, 0) \Vert \bm{z} \Vert^2 - 2z_i \sum_{j=1}^{d}[{\rm max}(z_j, 0)]^2}{(\sum_{j=1}^{d} z_j^2)^2}, \\
			\frac{\partial p_{L1}}{\partial z_i} &=
			\begin{dcases}
				\frac{\sum_{j=1}^{d} {\rm max}(z_j, 0)}{\Vert \bm{z} \Vert_1^2}, & z_i < 0 \\
				\frac{\Vert \bm{z} \Vert_1 - \sum_{j=1}^{d} {\rm max}(z_j, 0)}{\Vert \bm{z} \Vert_1^2}, & z_i > 0
			\end{dcases}, \\
			\frac{\partial p_c}{\partial z_i} &= 0.
		\end{aligned}
	\end{equation}

	Thus, we obtained all gradients of CAS for gradient descent and backpropagation. Since Eq. \eqref{CAReLU} and Eq. \eqref{CASBNReLU} are just different concatenations of CAS, ReLU, and BN, which are all differentiable with respect to the input and parameters, the gradients can be derived from the chain rule.

	\section{Experiments}
	In this section, we evaluate CAReLU across different applications. For models without BN, we directly replace existing activation functions with Eq. \eqref{CAReLU} in every layer. For models with BN, we also replace $\rm BN\text{-}ReLU$ combination with Eq. \eqref{CASBNReLU} in every layer. We respectively use $\rm CAReLU_E$, $\rm CAReLU_{L1}$, and $\rm CAReLU_c$ to denote CAReLU implemented with $p_E$ in Eq. \eqref{p_E}, $p_{L1}$ in Eq. \eqref{p_{L1}}, and $p_c$ in Eq. \eqref{p_cnt}. The following experiments show that our method outperforms other popular activation functions in multiple applications.

	\subsection{CIFAR-100 Image Classification}

	We compare our methods to the most popular activation functions on CIFAR-100 imagine classification task \cite{cifar}. CIFAR-100 is a dataset that has 100 classes, containing 500 training images and 100 test images for each class. We use ResNet-18 \cite{resnet}, GoogLeNet \cite{googlenet}, and VGG-13 \cite{vgg} networks to evaluate our activation functions. A stochastic gradient descent (SGD) optimizer with a momentum of $0.9$ and a weight decay of $5 \times 10^{-4}$ is used to train all networks. The learning rate starts at $0.1$ and is divided by $5$ in $50$th, $120$th, and $160$th epochs. The training ends at $200$th epoch.

	\begin{table}[ht]
		\setlength{\tabcolsep}{4pt}
		\centering
		\footnotesize
		\caption{Top-1 Accuracy (\%) on CIFAR-100 Test Set}
		\label{cifar-100_results}
		\begin{tabular}{ccccc}
			\toprule  
			Methods & ResNet-18 & GoogLeNet & VGG-13 \\
			\midrule  
			${\rm ReLU}$ & $76.14 \pm 0.21$ & $78.53 \pm 0.19$ & $72.53 \pm 0.20$ \\
			${\rm LeakyReLU}$ & $76.18 \pm 0.14$ & $78.43 \pm 0.08$ & $72.27 \pm 0.11$ \\
			${\rm PReLU}$ & $74.26 \pm 0.14$ & $76.14 \pm 0.37$ & $71.02 \pm 0.15$ \\
			${\rm Swish\text{-}1}$ & $75.74 \pm 0.15$ & $75.74 \pm 0.08$ & $71.36 \pm 0.07$ \\
			${\rm Swish}$ & $76.30 \pm 0.25$ & $77.31 \pm 0.24$ & $72.10 \pm 0.25$ \\
			${\rm BN\text{-}CAReLU_E}$ & $76.50 \pm 0.20$ & $\bm{79.23 \pm 0.13}$ & $\bm{72.85 \pm 0.19}$ \\
			${\rm CAReLU_E}$ & $\bm{76.62 \pm 0.23}$ & $79.21 \pm 0.15$ & $72.62 \pm 0.14$ \\
			${\rm BN\text{-}CAReLU_{L1}}$ & $76.44 \pm 0.15$ & $78.94 \pm 0.18$ & $72.47 \pm 0.32$ \\
			${\rm CAReLU_{L1}}$ & $76.43 \pm 0.15$ & $78.89 \pm 0.22$ & $72.58 \pm 0.36$ \\
			${\rm BN\text{-}CAReLU_c}$ & $76.29 \pm 0.28$ & $78.60 \pm 0.28$ & $72.46 \pm 0.23$ \\
			${\rm CAReLU_c}$ & $76.13 \pm 0.21$ & $78.65 \pm 0.11$ & $72.24 \pm 0.26$ \\
			\bottomrule 
		\end{tabular}
	\end{table}

	Table \ref{cifar-100_results} shows the results of the classification experiment. Most implementations of our method have overall better performance compared to other baseline methods. ${\rm CAReLU_E}$ has the highest accuracy in ResNet-18 experiments and ${\rm BN\text{-}CAReLU_E}$ has the highest accuracy in GoogLeNet and VGG-13 experiments. When comparing different implementations of our method, ${\rm BN\text{-}CAReLU}$ performs better than ${\rm CAReLU}$ except for the ${\rm CAReLU_E}$ in ResNet-18 experiments and ${\rm CAReLU_c}$ in GoogLeNet, which proves the previous assumption that BN's normalization impairs the effectiveness of our method. Energy-based implementations generally perform better than the other two implementations. The reason is that the gradient of ${\rm CAReLU_E}$ is a continuous function and it's smoother than the gradient of ${\rm CAReLU_{L1}}$; ${\rm CAReLU_c}$ contains Heaviside step functions, which results in a bumpy loss landscape \cite{landscape} and thus impairs the performance.

	\begin{table}[ht]
		\centering
		\caption{Values of $\tanh(\alpha p + \beta)$ from the First 10 CAReLUs of \\Best Trained Models}
		\label{cifar100_scale_factor}
		\begin{tabular}{ccccc}
			\toprule  
			$\rm CAS$ & ResNet-18 & GoogLeNet & VGG-13 \\
			\midrule   
			\#1 & $0.5418 \pm 0.0172$ & $0.1775 \pm 0.0007$ & $0.1861 \pm 0.0028$ \\
			\#2 & $0.1915 \pm 0.0113$ & $0.2028 \pm 0.0169$ & $0.1447 \pm 0.0078$ \\
			\#3 & $0.4540 \pm 0.0218$ & $0.1449 \pm 0.0049$ & $0.2670 \pm 0.0265$ \\
			\#4 & $0.0678 \pm 0.0065$ & $0.1591 \pm 0.0079$ & $0.1489 \pm 0.0077$ \\
			\#5 & $0.2266 \pm 0.0060$ & $0.1558 \pm 0.0250$ & $0.1388 \pm 0.0045$ \\
			\#6 & $0.1327 \pm 0.0100$ & $0.5597 \pm 0.0305$ & $0.9923 \pm 0.0000$ \\
			\#7 & $0.5954 \pm 0.0067$ & $0.3055 \pm 0.0020$ & $0.0962 \pm 0.0085$ \\
			\#8 & $0.1049 \pm 0.0046$ & $0.2143 \pm 0.0399$ & $0.4725 \pm 0.0060$ \\
			\#9 & $0.1795 \pm 0.0033$ & $0.1317 \pm 0.0080$ & $0.1553 \pm 0.0245$ \\
			\#10 & $0.1673 \pm 0.0149$ & $0.4171 \pm 0.0008$ & $0.2692 \pm 0.0849$ \\
			\bottomrule 
		\end{tabular}
	\end{table}



	\begin{figure*}[ht]
		\subfigure[]{\includegraphics[scale=0.60]{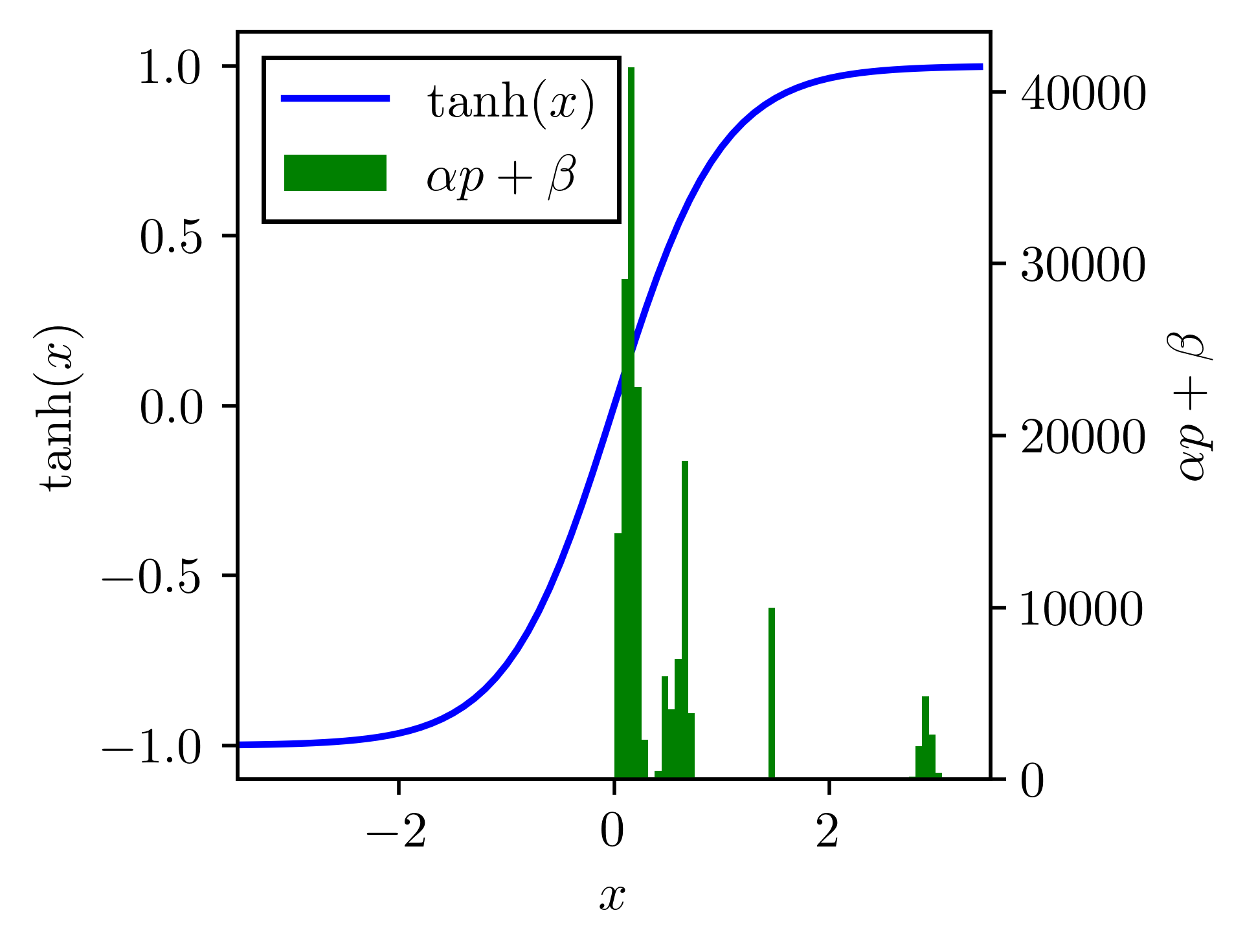}}
		\hfill
		\subfigure[]{\includegraphics[scale=0.60]{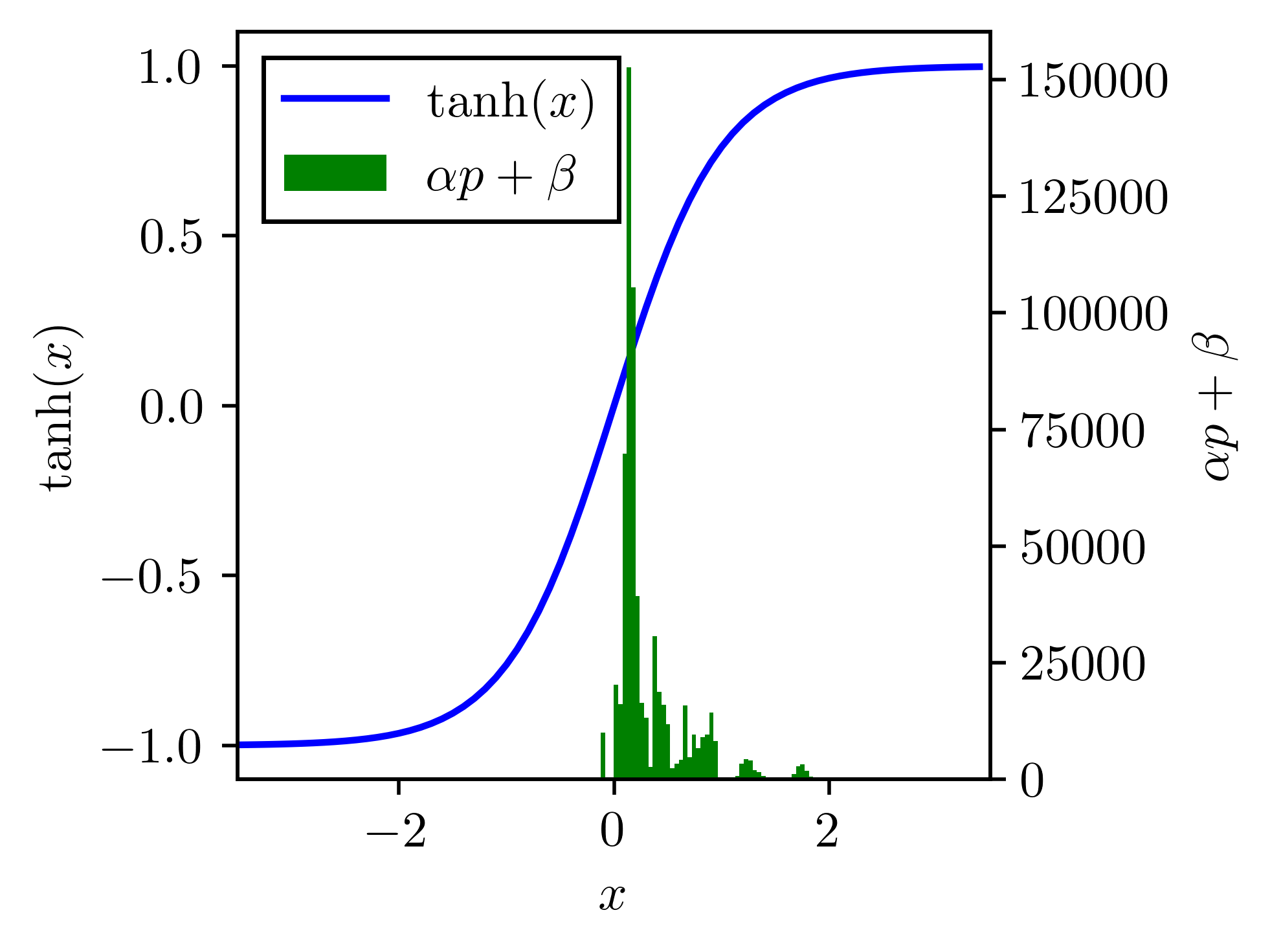}}
		\hfill
		\subfigure[]{\includegraphics[scale=0.60]{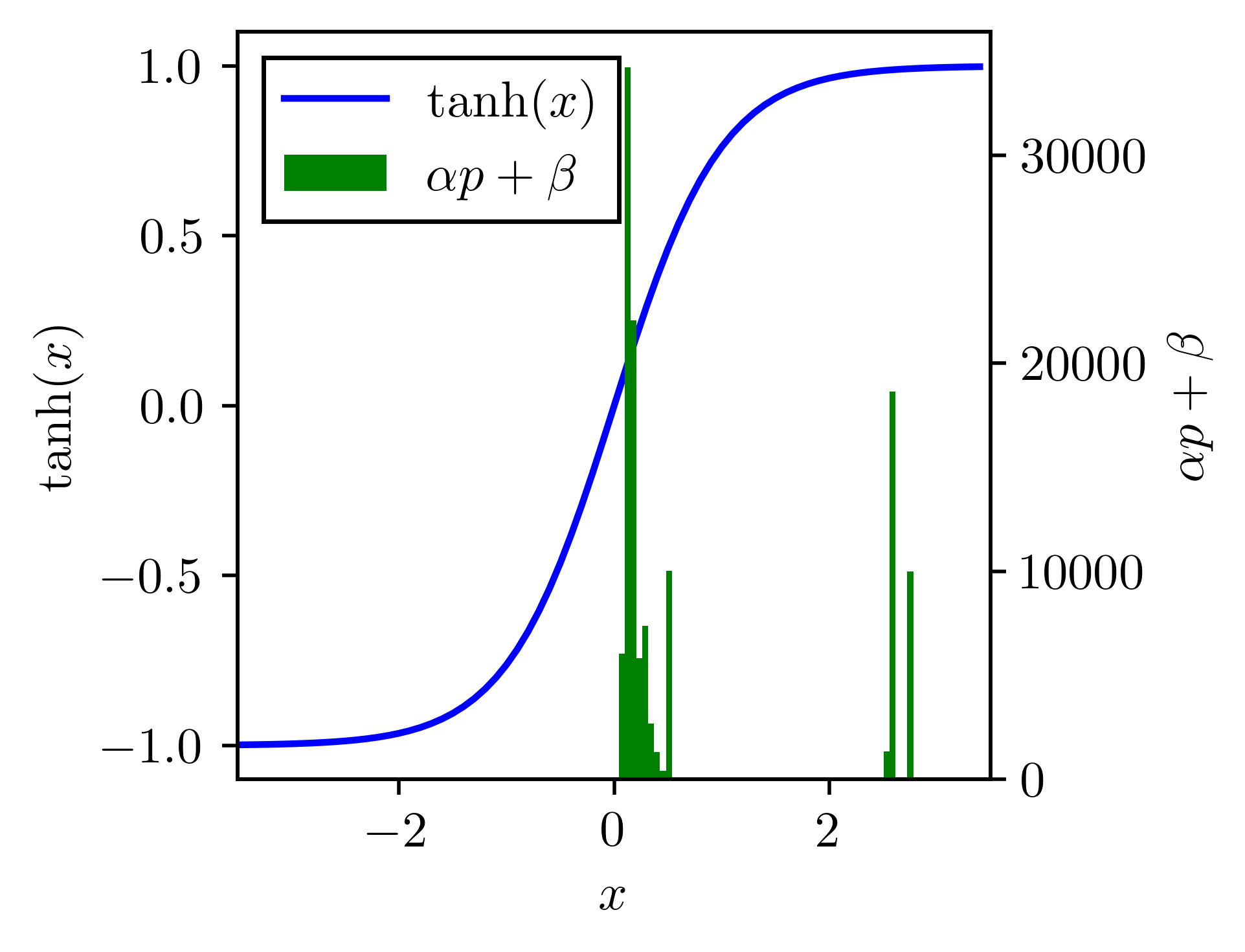}}
		\caption{Histograms of $\alpha p + \beta$ obtained from the best trained model. (a) ResNet-18/${\rm CAReLU_E}$. (b) GoogLeNet/${\rm BN\text{-}CAReLU_E}$. (c) VGG-13/${\rm BN\text{-}CAReLU_E}$.}
		\label{cifar_100_scalar_distribution}
	\end{figure*}
	
	We also evaluate values of $\tanh(\alpha p + \beta)$ in CASs on each layer. In Table \ref{cifar100_scale_factor}, we show the means and standard deviations of $\tanh(\alpha p + \beta)$ from the first 10 CASs of best models on the test set. Small deviations indicate that CASs on trained models do not generate significantly different values of $\tanh(\alpha p + \beta)$ for different samples. Instead, input values are scaled uniformly and then fine-tuned for each sample according to the input's competition results. Some degenerate into constant scaling such as CAS\#6 of VGG-13 in Table \ref{cifar100_scale_factor}. Allowing degeneracy into ReLU enables our method to utilize competition results without compromising the performance whenever the original approach is optimal. Histograms of $\alpha p + \beta$ obtained from the best models on the test set are shown in Fig. \ref{cifar_100_scalar_distribution}. Despite our initial design being a binary scale factor described in Eq. \eqref{SDF1}, values of $\alpha p + \beta$ mostly land on non-saturating regions.
	
	\subsection{BSD-300 Image Super Resolution}

	We compare our methods to the most popular activation functions on an image super-resolution task on the Berkeley segmentation dataset (BSD300), which contains 200 training images and 100 test images \cite{bsd}. The network we used for this experiment is an efficient sub-pixel convolutional neural network (ESPCN) featuring a network comprising several convolutional layers and a pixel shuffle layer \cite{espcn}. Training images are cropped to $256 \times 256$ and scaled down to $(256/r) \times (256/r)$ where $r$ is the upscale factor. The ESPCN network upscales the down-scaled images back to $256 \times 256$. An Adam optimizer \cite{adam} with a learning rate of $1 \times 10^{-3}$ is used for training the network in 200 epochs. We run every setting 5 times and show experiment results in Table \ref{bsd300_results}. 

	The data show that ${\rm CAReLU_E}$ and ${\rm CAReLU_{L1}}$ surpass other activation functions in PSNR. ${\rm CAReLU_c}$ outperforms PReLU, Swish-1, and Swish, but does not perform as well as ${\rm CAReLU_E}$ and ${\rm CAReLU_{L1}}$ due to its discontinuity introduced by the Heaviside function.

	\begin{table}[ht]
		\centering
		\footnotesize
		\caption{PSNR (dB) on BSD300 Super-resolution Task}
		\label{bsd300_results}
		\begin{tabular}{ccc}
			\toprule  
			Methods & $r=3$ & $r=4$ \\
			\midrule   
			${\rm ReLU}$ & $25.001 \pm 0.014$ & $23.634 \pm 0.008$ \\
			${\rm LeakyReLU}$ & $25.005 \pm 0.008$ & $23.635 \pm 0.010$ \\
			${\rm PReLU}$ & $24.980 \pm 0.015$ & $23.635 \pm 0.014$ \\
			${\rm Swish\text{-}1}$ & $24.934 \pm 0.010$ & $23.585 \pm 0.010$ \\
			${\rm Swish}$ & $24.947 \pm 0.013$ & $23.590 \pm 0.011$ \\
			${\rm CAReLU_E}$ & $ \bm{25.018 \pm 0.006} $ & $23.642 \pm 0.008$ \\
			${\rm CAReLU_{L1}}$ & $25.015 \pm 0.003$ & $ \bm{23.643 \pm 0.009} $ \\
			${\rm CAReLU_c}$ & $25.000 \pm 0.012$ & $23.631 \pm 0.013$ \\
			\bottomrule 
		\end{tabular}
	\end{table}

	\subsection{Natural Language Inference on SNLI}

	In this section, we evaluate our methods on the Stanford Natural Language Inference (SNLI) corpus \cite{snli}. SNLI corpus is a collection of 570k human-written English sentence pairs labeled for entailment, contradiction, and neutral. A premise and a hypothesis comprise a sentence pair. The model we used for this task comprises an embedding layer, a Long short-term memory (LSTM) \cite{lstm} encoder, and a sequence of fully connected layers. Hypothesis and premise go through the embedding layer and the encoder independently. We concatenate the encoded hypothesis feature and premise feature, then send them through the fully connected layers for classification. We use the Adam optimizer with a learning rate of $0.001$ to train the parameters for 50 epochs on the SNLI training set. We run every setting 5 times and show experiment results in Table \ref{snli_result}. 

	\begin{table}
		\centering
		\begin{threeparttable}[ht]
			\footnotesize
			\setlength{\tabcolsep}{30pt}
			\caption{Classification Accuracy (\%) on SNLI Test Set}
			\label{snli_result}
			\begin{tabular}{cc}
				\toprule  
				Methods & Acc \\
				\midrule  
				${\rm ReLU}$ & $77.69 \pm 0.25$ \\
				${\rm LeakyReLU}$ & $77.12 \pm 0.48$ \\
				${\rm PReLU}$ & $77.25 \pm 0.33$ \\
				${\rm Swish\text{-}1}$ & $-$\tnote{1} \\
				${\rm Swish}$ & $-$ \\
				${\rm CAReLU_E}$ & $\bm{78.67 \pm 0.25}$ \\
				${\rm CAReLU_{L1}}$ & $78.41 \pm 0.39$ \\
				${\rm CAReLU_c}$ & $78.40 \pm 0.25$ \\
				\bottomrule 
			\end{tabular}
			\begin{tablenotes}
				\item [1] "$-$" indicates that the training with this activation function does not converge.
			\end{tablenotes}
		\end{threeparttable}
	\end{table}
	
	Data suggest that our methods perform better than other activation functions. Swish and Swish-1 are unable to converge in this task. ${\rm CAReLU_E}$ achieve the highest classification accuracy in this experiment. Though mean accuracies of ${\rm CAReLU_{L1}}$ and ${\rm CAReLU_c}$ are approximately identical, ${\rm CAReLU_c}$ is more robust because it has a smaller standard deviation. All three implementations of our method outperform other baseline activation functions.
	

	\section{Conclusion}
	Stemming from the idea that both positive values and negative values are equally important and they must compete for activation, we developed the Competition-based Adaptive ReLU (CAReLU) activation function by introducing the competition between positive values and negative values. A CAReLU has 2 parameters that can be trained uniformly with other network parameters using gradient descent. By respectively implementing CAReLU with each one of 3 competition indicators, we developed 3 different activation functions: $\rm CAReLU_E$, $\rm CAReLU_{L1}$, and $\rm CAReLU_c$. We also developed a technique when working with Batch Normalization to have extra performance gain. 
	
	We evaluated our method on different tasks and achieved consistent performance improvements. $\rm CAReLU_E$ is generally more effective, but $\rm CAReLU_{L1}$ and $\rm CAReLU_c$ can also achieve great performance in certain scenarios. The effectiveness and the new perspective on the competition between positive values and negative values make CAReLU promising in deep learning tasks.

	\bibliography{cite}

\end{document}